\documentclass{article}

\usepackage[final, nonatbib]{neurips_2024}

\usepackage[utf8]{inputenc} % allow utf-8 input
\usepackage[T1]{fontenc}    % use 8-bit T1 fonts
\usepackage{hyperref}       % hyperlinks
\usepackage{url}            % simple URL typesetting
\usepackage{booktabs}       % professional-quality tables
\usepackage{amsfonts}       % blackboard math symbols
\usepackage{nicefrac}       % compact symbols for 1/2, etc.
\usepackage{microtype}      % microtypography
\usepackage{xcolor}         % colors

\usepackage{graphicx}
\usepackage{wrapfig}

\usepackage{amsmath}
\usepackage{amssymb}
\usepackage{environ}
\usepackage{slashed}
\usepackage{mathrsfs}
\usepackage{subcaption}

\usepackage{algorithm}
\usepackage{algpseudocode}

\usepackage{footnote}

\usepackage{multirow}
\usepackage{array}
\newcolumntype{C}[1]{>{\centering\let\newline\\\arraybackslash\hspace{0pt}}m{#1}}
\newcolumntype{M}[1]{>{\centering\arraybackslash}m{#1}}

\NewEnviron{derivation}
{%
    \\
    \begin{minipage}{\linewidth}
        \begin{equation}\begin{split}
        \BODY
        \end{split}\end{equation}
    \end{minipage}
}

\makeatletter
\newcommand{\mfootnote}[1]{%
  \ifmeasuring@
    \chardef\@tempfn=\value{footnote}%
    \! \footnotemark
    \setcounter{footnote}{\@tempfn}%
  \else
    \iffirstchoice@
      \! \footnote{#1}%
    \fi
  \fi}
\makeatother
\graphicspath{ {images/} }
\title{Mixture of Scales: Memory-Efficient Token-Adaptive Binarization for Large Language Models}

\author{
  Dongwon Jo$^{\,1}$\quad Taesu Kim$^{\,2}$\quad Yulhwa Kim$^{\,3\,\thanks{Corresponding Author}}$\quad Jae-Joon Kim$^{\,1\,\footnotemark[1]}$\\
  $^{1\,}$Seoul National University\quad$^{2\,}$SqueezeBits Inc.\quad$^{3\,}$Sungkyunkwan University\\
  \texttt{\{dongwonjo, kimjaejoon\}@snu.ac.kr}\\
  \texttt{\{taesu.kim\}@squeezebits.com}\\
  \texttt{\{yulhwakim\}@skku.edu}
}

\begin{document}

\maketitle

\begin{abstract}
% Though large language models (LLMs) have delivered impressive results in a variety of natural language processing (NLP) tasks, their massive size often complicates deployment. 
Binarization, which converts weight parameters to binary values, has emerged as an effective strategy to reduce the size of large language models (LLMs). However, typical binarization techniques significantly diminish linguistic effectiveness of LLMs.
To address this issue, we introduce a novel binarization technique called Mixture of Scales (BinaryMoS). Unlike conventional methods, BinaryMoS employs multiple scaling experts for binary weights, dynamically merging these experts for each token to adaptively generate scaling factors. This token-adaptive approach boosts the representational power of binarized LLMs by enabling contextual adjustments to the values of binary weights. Moreover, because this adaptive process only involves the scaling factors rather than the entire weight matrix, BinaryMoS maintains compression efficiency similar to traditional static binarization methods. Our experimental results reveal that BinaryMoS surpasses conventional binarization techniques in various natural language processing tasks and even outperforms 2-bit quantization methods, all while maintaining similar model size to static binarization techniques.
\end{abstract}

\section{Introduction}
Though large language models (LLMs) have delivered impressive results in a variety of natural language processing (NLP) tasks, their massive size often complicates deployment.
One common method to compress LLMs is through the quantization of weight parameters, which reduces model sizes by lowering the precision of weight values~\cite{llm_quant, model_compression, gptq, qlora, spqr, sqzllm, quip, awq, owq}. Existing quantization approaches such as GPTQ~\cite{gptq}, AWQ~\cite{awq}, and OWQ~\cite{owq} have successfully managed to reduce model sizes by converting 16-bit floating point weights to 4-bit representations, achieving a fourfold decrease in size. Binarization pushes this concept even further by reducing weight values to 1-bit, resulting in a 16-fold size reduction. 

However, such aggressive compression through binarization drastically limits the representational capacity of weights, leading to a significant degradation in the linguistic capabilities of LLMs. 
To address this limitation and improve the accuracy of binarized LLMs, recent research has actively explored binarization techniques tailored for LLMs~\cite{bitnet, pbllm, billm, onebit}. Nonetheless, previous efforts often compromise the inherent advantages of binarization by introducing high memory overhead, and they continue to struggle to achieve sufficient accuracy with binarized LLMs.

In this paper, we propose a novel binarization technique named as Mixture of Scales (BinaryMoS).
Typical binarization methods use scaling factors to control the effective values of binarized weights. Although these scaling factors occupy a tiny fraction of the overall model size, they are crucial in reducing binarization error.
BinaryMoS advances the functionality of these scaling factors by incorporating token-adaptive scaling factors.
Inspired by the Mixture of Experts (MoE) approach~\cite{moe, switch_trans, stmoe}, which empolys multiple expert layers to enhance the model capacity, BinaryMoS adopts multiple scaling factors as experts to improve the representational capacity of binarized LLMs in a memory-efficient way.
During inference, BinaryMoS linearly combines these scaling experts based on the context to generate token-adaptive scaling factors, thus dynamically adjusting the represented values of binarized weights to maximize the expressive power of the model.
As a result, BinaryMoS can improve the linguistic performance of binarized LLMs with minimal memory overhead.

\section{Background}

\begin{figure}[t]
    \centering
    \includegraphics[width=0.85\linewidth]{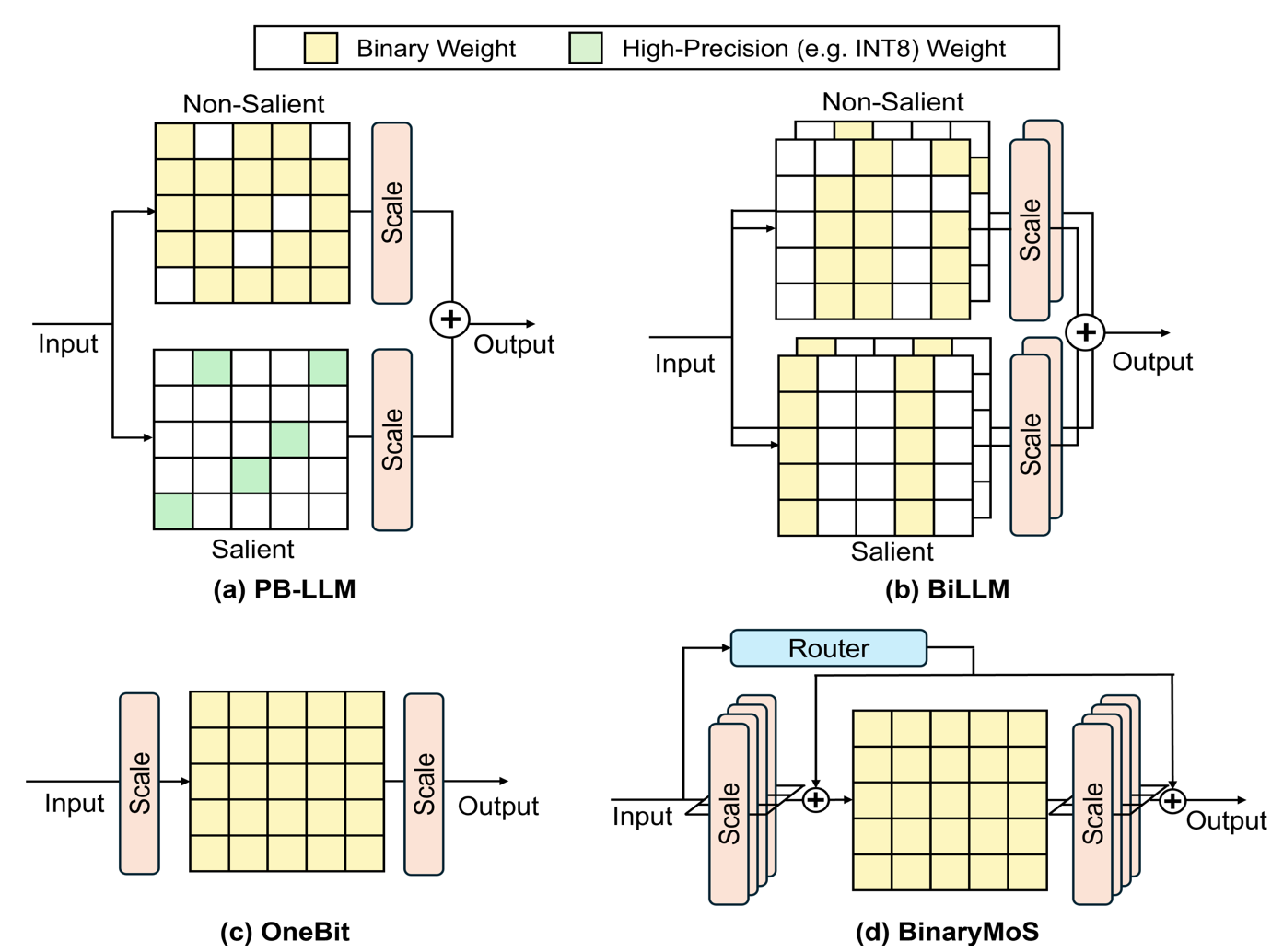}
    % \vspace{-12pt}
     \vspace{-6pt}
    \caption{
        A brief overview of various LLM binarization methods. 
        PB-LLM involves both a binary weight matrix and a high-precision, sparse weight matrix, and BiLLM stores four types of binary weight matrices. 
        OneBit simplifies the layer structure by introducing scaling factors for input and output dimensions respectively.
        BinaryMoS introduces multiple scaling experts to enhance the capacity of binarized models.
    }
    \vspace{-12pt}
    \label{fig:binarization}
\end{figure}

\subsection{Binarization of LLMs} \label{sec:previous_binarziation}
Binarization stands out as an extreme yet effective method for reducing model sizes in deep learning. This method achieves size reduction by transforming high-precision weight parameters into 1-bit values.
The binarization process is typically governed by the following equation:
\begin{equation}
% \begin{aligned}
    \label{eq:conventional_binarization}
    W_{B} = \alpha \cdot \text{Sign}(W_{FP}- \overline{W}_{FP})
% \end{aligned}
\end{equation}
Here, $W_{FP} \in \mathbb{R}^{n \times m}$ is the full-precision weight matrix of a linear layer where $n$ and $m$ represent the size of output and input dimension, respectively, and $W_{B} \in \mathbb{R}^{n \times m}$ denotes its binarized version. 
$\alpha \in \mathbb{R}^{n}$ represents scaling factors that are responsible for adjusting the binary weight values along the output dimension.
In general, the scaling factors are analytically derived as the absolute mean of FP weight values to minimize the $L2$ error between full-precision and binarized weights, and these scaling factors play a vital role in bridging the gap between the original full-precision weights and their binarized counterparts.

While binarization has been effectively applied in traditional deep learning models like Convolutional Neural Networks (CNNs) for image classification without losing accuracy~\cite{binary_survey, xnor, bireal, react}, LLMs tend to be more sensitive to such extreme quantization, often experiencing significant accuracy degradation with standard binarization techniques.
Therefore, various binarziation techniques tailored for LLMs have been developed, as shown in Figure~\ref{fig:binarization}. 
% To address this, 
PB-LLM~\cite{pbllm} partially binarizes weight parameters while maintaining salient weight parameters as high-precision values (e.g., Float16 or INT8). However, this method results in considerable memory overhead. For instance, quantizing 10\% of weight parameters as INT8 while binarizing the remaining 90\% results in an average bit-width of 1.7 bits for the weight parameters, which is closer to 2 bits than 1 bit.

Furthermore, despite this partial binarization strategy of PB-LLM, the significant information loss inherent in binarization still causes considerable accuracy degradation.
To reduce the binarization error and enhance accuracy, BiLLM~\cite{billm} adopts a more refined approach to assigning scaling factors. 
Assuming that weight parameters follow a bell-shaped distribution, BiLLM categorizes weight parameters based on their proximity to the mean value: concentrated weights, close to the mean, and sparse weights, distant from the mean. Distinct scaling factors are then assigned to each group to minimize binarization errors. Then, to reduce the memory overhead associated with maintaining information of salient weights, BiLLM preserves this information by binarizing the difference between the binarized values and their full-precision counterparts. Consequently, each salient weight is represented by two 1-bit values, effectively amounting to a 2-bit representation.
Despite significantly reducing binarization error, BiLLM complicates the structure of binarized LLMs, adding complexity to the inference process. 
This complexity arises from the need to manage additional sparse and salient weights alongside regular concentrated weights, requiring extra matrix multiplication during inference.

Meanwhile, unlike conventional binarization methods that typically employ scaling factors only for the output dimension of weights, OneBit~\cite{onebit} enhances the binarization process by incorporating scaling factors for both the input and output dimensions.
This dual-dimension scaling approach addresses binarization errors across both dimensions, potentially enhancing model accuracy.
Additionally, the size of each scaling vector is substantially smaller compared to the weight matrix, making this approach memory efficient. For instance, in linear layers with a hidden dimension of $h$, the weight matrix size is $h \times h$, while each scaling vector is only $h \times 1$. Therefore, doubling these scaling factors adds a negligible memory overhead to the network.
Moreover, as this approach of dual-dimensional scaling efficiently preserves enough information to significantly reduce binarization errors, OneBit eliminates the need to store separate information for salient weights, thereby simplifying the model structure.
The result of matrix multiplication $Y$ of a linear layer using the OneBit approach can be defined as follows:
\begin{equation}
% \begin{aligned}
    \label{eq:svd_binarization}
    Y = X [S_{in}^T \odot \text{Sign}(W^T_{FP}) \odot S_{out}] = [(X \odot S_{in})\text{Sign}(W^T_{FP})] \odot S_{out}
% \end{aligned}
\end{equation}
Here, $X \in \mathbb{R}^{k \times m}$ is the matrix of input activation where $k$ represents batch size, while $S_{in}\in \mathbb{R}^{1 \times m}$ and $S_{out}\in \mathbb{R}^{1 \times n}$ denote the scaling factors for input and output dimensions, respectively.
As outlined in Equation~\ref{eq:svd_binarization}, processing scaling factors for both input and output dimension can be simplified to scaling input and output of the linear layer before and after matrix multiplication, respectively.

Despite advances in binarization techniques for LLMs, a notable accuracy gap still exists between full-precision models and their binarized counterparts. Therefore, bridging this gap without sacrificing the fundamental benefits of binarization, particularly low memory usage, remains an important challenge in the field of LLM compression.

\subsection{Mixture of Experts}
The MoE approach is a widely adopted strategy to boost the capabilites of deep learning models by integrating multiple specialized experts into a single framework~\cite{moe, switch_trans, stmoe}. Typically, the MoE approach for LLMs involves duplication of layers and selecting the appropriate layers among these duplicates for a specific task during inference.
In the MoE setup, the router is a key to selecting the appropriate expert. It generally consists of a linear layer followed by a softmax function, which calculates and assigns scores to each expert. During the inference, only the experts with the highest score are selected and processed.

While integrating the MoE approach with binarized LLMs offers potential for improving model accuracy, it presents a substantial memory trade-off. The duplication of layers inherent in MoE increases the model size proportionally with the number of experts, thus diminishing the memory efficiency benefits gained from binarization.
To address these challenges, we propose BinaryMoS, a novel binarization technique that aims to enhance model capacity while maintaining memory efficiency. This approach leverages scaling factors as experts, improving accuracy of binarized LLMs without the extensive memory overhead associated with traditional MoE configurations. In the following section, we will delve deeper into how BinaryMoS operates and its benefits over conventional techniques.

\section{Proposed BinaryMoS}

\begin{figure}[t]
    \centering
    \includegraphics[width=1.0\linewidth]{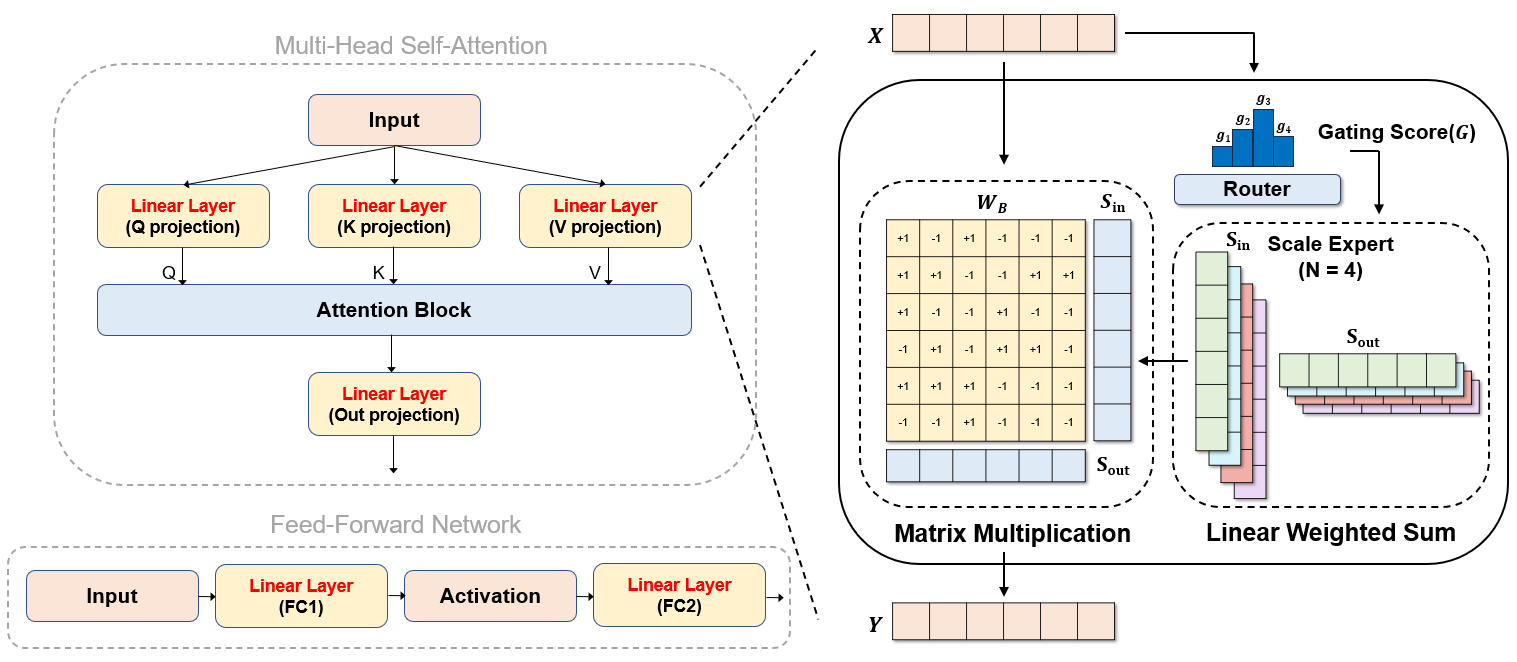}
    % \vspace{-10pt}
        \vspace{-13pt}
    \caption{
        Illustration of the proposed BinaryMoS scheme. The proposed BinaryMoS introduce mixture of scale approach to generate token-adaptive scaling factors.
    }
    \vspace{-12pt}
    \label{fig:BinaryMoS}
\end{figure}

\subsection{Binarization with Mixture of Scale}
An overview of the proposed BinaryMoS is presented in Figure~\ref{fig:BinaryMoS}.
Unlike previous binarization techniques that utilize a single scaling vector per input or output dimension, BinaryMoS integrates the concept of experts from the MoE framework into the scaling factors and utilizes multiple scaling experts for each dimension.
As discussed in Section~\ref{sec:previous_binarziation}, although the size of scaling factors is relatively small, they play a crucial role in preserving the accuracy of binarized models. Therefore, introducing multiple scaling experts incurs minimal memory overhead while effectively leveraging the advantages of the MoE strategy to enhance the capabilities of binarized models.

In the MoE framework, the number of experts selected corresponds directly to the number of layers processed. As a result, the typical MoE framework selects only one or two experts per inference stage to manage the increased processing burden associated with more experts being selected.
On the other hand, the scaling factors of binarized LLMs are solely involved in linear operations with matrix multiplication, as detailed in Equation~\ref{eq:svd_binarization}.
This linearity allows for the efficient management of multiple scaling experts by linearly combining them before executing the matrix multiplication.
Hence, instead of selecting only a few experts, as done in the conventional MoE framework, BinaryMoS dynamically generates instructions on how to combine these scaling experts based on the context. This approach overcomes the limitations of fixed expert choices in typical MoE setups by enabling the creation of effectively infinite token-adaptive scaling factors through linear combinations.
Consequently, by optimally utilizing the representational power of multiple scaling experts, BinaryMoS maximizes the potential of binarized models while maintaining memory efficiency.

\subsection{Router Design}
In order to generate the token-adaptive scaling factors, the proposed BinaryMoS designs the router for processing the following operations:
\begin{equation}
    G = \text{Softmax}(XW_{R})
    \label{eq:BinaryMoS1}
\end{equation}
\begin{equation}
    \hat{S}_{in} = GS_{in},\quad\hat{S}_{out} = GS_{out}
    \label{eq:BinaryMoS2}
\end{equation}
Here, $W_{R} \in \mathbb{R}^{m \times e}$ represents the weight parameters of router's linear layer, where $e$ denotes the number of experts. $S_{in} \in \mathbb{R}^{e \times m}$ and $S_{out} \in \mathbb{R}^{e \times n}$ denote the scaling experts for input and output dimension, respectively.
Initially, the router computes the gating score $G$, which represents the significance of each scaling expert, using input activations and router weights, as outlined in Equation~\ref{eq:BinaryMoS1}.
Notably, as the gating scores are generated with the softmax function, the sum of gating scores for the scaling experts equals 1.
These scores are used to linearly combine the scaling experts, resulting in the creation of token-adaptive scaling factors $\hat{S}_{in}$ and $\hat{S}_{out}$, as shown in Equation~\ref{eq:BinaryMoS2}.
Then, by replacing the static scaling factors $S_{in}$ and $S_{out}$ from Equation~\ref{eq:svd_binarization} with token-adaptive scaling factors, the result of matrix multiplication $\hat{Y}$ in a linear layer using the BinaryMoS approach can be revised as follows:
%matrix multiplication for a output $Y$ of linear layer and using the BinaryMoS approach can be revised as follows:
\begin{equation}
\label{eq:BinaryMoS3}
\hat{Y} = [(X \odot \hat{S}_{in})\text{Sign}(W^T_{FP})] \odot \hat{S}_{out}
\end{equation}
We empirically find that using four scaling experts each for the input and output dimensions provides the optimal compromise between increasing model size and improving accuracy.
Consequently, the proposed BinaryMoS utilizes four scaling experts for each dimension to enhance accuracy while maintaining efficiency.

\subsection{Impact of BinaryMos on LLM Compression}
\begin{table}[t]
\setlength{\tabcolsep}{3.5pt}
  \centering
    \caption{Comparison of memory requirements for deploying Float16 and binarized models, with the number in parentheses denoting the compression ratio of binarized models over Float16 models.}
  \label{table:memory}
  \vspace{6pt}
  \renewcommand{\arraystretch}{1.20}
  % \footnotesize
  \scalebox{0.85}{
    \begin{tabular}{l|c |c | c | c| c}
    \toprule
    Model & Float16 & PB-LLM & BiLLM & OneBit & BinaryMoS\\
    \midrule
    LLaMA-1/2-7B & 13.51 GB & 2.78 GB (4.86$\times$)& 2.28 GB (5.93$\times$)& 1.37 GB ( 9.86$\times$)& 1.40 GB ( 9.65$\times$)\\
    LLaMA-1/2-13B & 26.20 GB & 5.02 GB (5.22$\times$)& 4.06 GB (6.45$\times$)& 2.29 GB (11.44$\times$)& 2.33 GB (11.24$\times$)\\
    \bottomrule
  \end{tabular}
  }
  % \vspace{-6pt}
    \vspace{-10pt}
\end{table}

The proposed BinaryMoS introduces additional memory overhead due to multiple scaling experts and the weights of the router. However, this overhead is relatively minor. For instance, in the LLaMA-1/2-7B model~\cite{llama} with a hidden dimension $h$ of 4096, the weight matrix for the linear layers is typically 4096$\times$4096. If BinaryMoS adopts 4 scaling experts, this translates to four $\alpha$'s, each of dimension 4096$\times$1, for both input and output dimensions. Additionally, the weights of the router would be 4096$\times$4. 
Compared to the previous OneBit method, which requires a single $\alpha$ for both input and output dimensions, the additional components in BinaryMoS total 4096$\times$10 parameters. The number of these extra parameters constitutes only 0.2\% of the original weight parameters.

For a comprehensive examination of the impact of various binarization techniques, including BinaryMoS, on LLM compression, we evaluate the memory requirements of LLaMA models with Float16 parameters and after applying different binarization methods, as detailed in Table~\ref{table:memory}. Following standard practice, all binarization techniques exclude the embedding layer and lm-head from binarization.
Our analysis reveals that BinaryMoS significantly reduces the memory footprint of models, achieving compression ratios ranging from 9.65$\times$ to 11.24$\times$. As model size increases, the relative impact of additional parameters diminishes and the proportion of the unbinarized part decreases. Hence, we can achieve higher compression ratios for larger models.
For instance, the original LLaMA-1/2-13B model, requiring 26.20 GB for deployment, is impractical for edge devices due to its size. However, BinaryMoS reduces this model to just 2.33 GB, representing an 11.24-fold decrease in memory requirements. This significant reduction facilitates deployment on edge devices with typically limited memory capacities of 4 GB.

In contrast, PB-LLM and BiLLM methods achieve relatively lower compression ratios of around 5$\times$ and 6$\times$, respectively. This is primarily due to two reasons: first, PB-LLM and BiLLM methods must retain salient weight information, increasing the average bitwidth of weight parameters. Second, the handling of sparse weight matrices in these methods introduces overhead in indexing sparse weight matrices, limiting the achievable compression ratio.
OneBit achieves the highest compression ratio by only introducing dual-dimension scaling factors. Remarkably, BinaryMoS achieves a comparable compression ratio to OneBit, despite incorporating additional components for scaling experts. While the memory requirement of binarized models with BinaryMoS increases by only 2\% compared to OneBit, the inclusion of scaling experts offers much greater potential to significantly improve perplexity.

This analysis demonstrates that although BinaryMoS introduces additional parameters, the relative increase in memory requirement is modest. This makes BinaryMoS a viable option for enhancing accurcy of binarized models without imposing a significant memory burden.

\subsection{Quantization-Aware Knowledge Distillation}
Following training strategies adopted for network compression~\cite{llm_qat, ternarybert}, we adopt the knowledge distillation (KD) to transfer the knowledge of a full-precision teacher model to a binarized student model.
We employ the cross entropy (CE) loss to distill the logit knowledge.
This is calculated using the following equation:
\begin{equation}
    \label{eq:KD_loss}
    \mathcal{L}_{C E}=-\frac{1}{n} \sum_c \sum_{i=1}^n p_c^{\mathcal{T}}\left(X_i\right) \log \left(p_c^{\mathcal{S}}\left(X_i\right)\right)
\end{equation}
Here, $\mathcal{S}$ and $\mathcal{T}$ represent the student and teacher models respectively. $n$ denotes batch size, and $c$ is the number of classes.
Additionally, to minimize the distributional discrepancies in layer outputs, we incorporate a mean-squared error (MSE) based layer-to-layer (L2L) loss as follows:
\begin{equation}
    \label{eq:KD_total1}
    \mathcal{L}_{L2L}=\sum_{l=1}^{L} \operatorname{MSE}\left(\mathbf{H}_l^\mathcal{T}, \mathbf{H}_l^\mathcal{S}\right)
\end{equation}
In this loss, $\mathbf{H}_l^\mathcal{T}$ and $\mathbf{H}_l^\mathcal{S}$ are the output logits from the $l$-th layer of the teacher and student models, respectively.
The total loss function, integrating both CE and L2L distillation losses, is defined as:
\begin{equation}
\label{eq:KD_total2}
\mathcal{L} = \mathcal{L}_{CE} + \alpha \cdot \mathcal{L}_{L2L}
\end{equation}
where $\alpha$ is a hyperparameter that balances the contributions of the CE and L2L losses.
For the training of BinaryMoS, we empirically set $\alpha = 10$.

\section{Experiments}

\subsection{Experimental Settings}
\textbf{Models and Evaluation Datasets.} In our study, we evaluate BinaryMoS on various models, including those from the LLaMA-1~\cite{llama}, LLaMA-2~\cite{llama2}, and OPT~\cite{opt} families. Specifically, we utilize the OPT models with 125M and 1.3B parameters, and the LLaMA-1 and LLaMA-2 models with 7B and 13B parameters for our evaluations.
We measure language modeling capabilities of these models by evaluating their perplexity on the WikiText2~\cite{wiki2} and C4~\cite{c4} datasets.
Additionally, we assess zero-shot accuracy on various Common Sense Reasoning Tasks such as BoolQ~\cite{boolq}, PIQA~\cite{piqa}, HellaSwag~\cite{hellaswag}, WinoGrande~\cite{winogrande}, ARC-e, ARC-c~\cite{arc}), utilizing the open-source LLM evaluation framework, LM-Evaluation-Harness~\cite{lm_eval}.

\textbf{Training Details.} 
We initialize the parameters of binarized models using those from pre-trained models, which serve as teacher models for KD. For the training dataset, a mixed dataset composed of the WikiText2 training dataset and a selected partition from the C4 training dataset, with a sequence length of 2048.
The training is conducted over three epochs using the AdamW~\cite{adamw} optimizer, with hyperparameters set to $\beta_{1}=0.9$, $\beta_{2}=0.999$, and zero weight decay. We implement a cosine decay learning rate scheduler, preceded by a warm-up phase constituting 0.03 of the total training duration. All training sessions are conducted on NVIDIA A100 GPUs.

\textbf{Baselines.}
We compare BinaryMoS against previous LLM binarization methods, including PB-LLM~\cite{pbllm}, BiLLM~\cite{billm}, and OneBit~\cite{onebit}, ensuring that all implementations adhere to the details provided in their respective papers. PB-LLM and BiLLM utilize the Post-Training Quantization (PTQ) approach for model calibration through the Optimal Brain Quantizer (OBQ) based method of GPTQ~\cite{gptq}. 
For PB-LLM, which allows variable ratios of salient weights to enhance accuracy, we have set the ratio of salient weights to 10\% to ensure the average bit width of weight parameters remains below 2 bits.
OneBit employs a Quantization-Aware Training (QAT) approach, and for fairness, its training setup is aligned with that of BinaryMoS. 
Given the significant accuracy improvements demonstrated by BinaryMoS over traditional binarization techniques, we also include a comparison with 2-bit quantization methods with PTQ approach, such as GPTQ~\cite{gptq} and OmniQuant~\cite{omniquant}, to broaden the evaluation scope.

 \subsection{Analysis on the Number of Scaling Experts} \label{sec:num_scale}
\begin{table}[t]
\setlength{\tabcolsep}{3.5pt}
  \centering
    \caption{The impact of the numbers of scaling experts on the proposed BinaryMoS. Quick assessment conducted using the LLaMA-1-7B model trained on one-third of the training data.}
  \label{table:num_scale}
  \vspace{6pt}
  \renewcommand{\arraystretch}{1.20}
  \scalebox{0.9}{
    \begin{tabular}{c|cc|cccccc|c}
    \toprule
    \multirow{2}{*}{\textbf{\# of Experts}} & \multicolumn{2}{c|}{\textbf{Perplexity $\downarrow$}} & \multicolumn{7}{c}{\textbf{Zero-shot Accuracy $\uparrow$}} \\ 
    & \textbf{Wiki2} & \textbf{C4} & BoolQ & PIQA & Hella. & WinoG. & ARC-e & ARC-c & \textbf{Average} \\ 
    \midrule
    \midrule
    1 & 9.33 & 12.54 & 60.27 & 67.84 & 46.77 & 52.09 & 38.38 & 27.98 & 48.89 \\
    2 & 9.19 & 12.18 & 62.69 & 68.55 & 48.36 & 55.09 & 40.23 & 28.92 & 50.64 \\
    4 & 8.92 & 11.85 & 60.51 & 67.46 & 49.95 & 55.24 & 41.16 & 29.35 & 50.61 \\
    8 & 9.17 & 12.28 & 58.68 & 67.46 & 47.51 & 53.67 & 39.52 & 29.43 & 49.38 \\
    \bottomrule
  \end{tabular}}
  \vspace{-8pt}
  % \vspace{-6pt}
\end{table}
To determine the optimal number of scaling experts for BinaryMoS, which effectively maintains the accuracy of binarized LLMs while minimizing memory usage, we conduct evaluations with LLaMA-1-7B using varying numbers of scaling experts. This evaluation is conducted using only one-third of the training data for quick assessment. As shown in Table~\ref{table:num_scale}, performance metrics, including perplexity and accuracy, generally improve as the number of experts increases from 1 to 4. However, a further increase to 8 experts leads to a decline in model performance. This decline arises from the challenge of training routers to appropriately assign scales to tokens as the number of scales increases. Based on these observations, we choose to employ 4 experts in the BinaryMoS approach.

\subsection{Analysis on the Token-Adaptive Scaling Factors}

\begin{figure}
    \centering
    \includegraphics[width=1.0\linewidth]{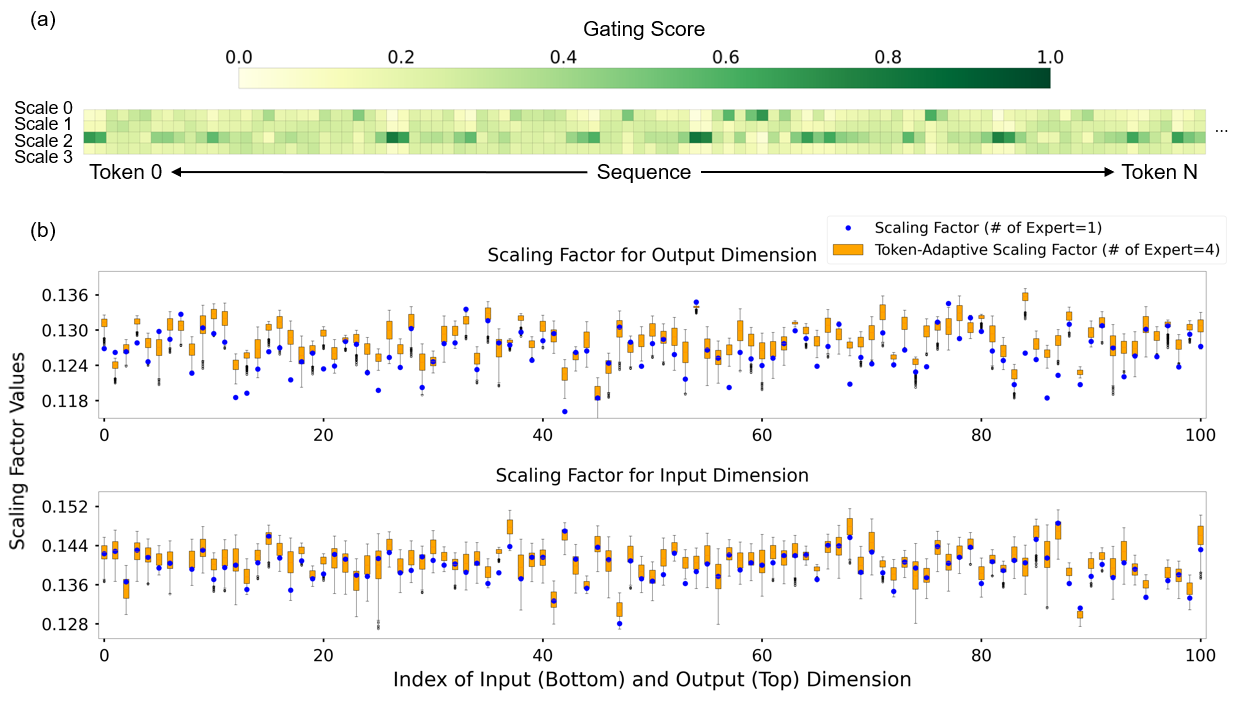}
    % \vspace{-12pt}
        \vspace{-15pt}
    \caption{
        ($\textbf{a}$) Gating scores of 4 scaling experts in 18th layer of LLaMA-1-7B model for each token in the input sequence. ($\textbf{b}$) Distribution of values of token-adaptive scaling factors.
        The boxplot visually presents the distribution of token-adaptive scaling factors among processed tokens. The box spans the interquartile range, indicating the middle 50\% of the scaling factors. Extending from the box are whiskers that reach the furthest data points within 1.5 times the interquartile range, providing insight into the overall range of the data.
    }
    \vspace{-12pt}
    \label{fig:routing_analysis}
\end{figure}

\begin{table}[t]
\setlength{\tabcolsep}{3.8pt}
  \centering
  \caption{Perplexity and zero-shot accuracy results of Float16 and binarized LLMs.}
  \label{table:binarized_rsults}
  \vspace{6pt}
  \renewcommand{\arraystretch}{1.27}
  % \scalebox{.75}{
  \scalebox{.81}{
    \begin{tabular}{llc|cc|cccccc|c}
    \specialrule{2.0pt}{-0.5pt}{0pt}
    %\toprule
    \multirow{2}{*}{\textbf{Model}} & \multirow{2}{*}{\textbf{Method}} & \multirow{2}{*}{\textbf{Wbits}} & \multicolumn{2}{c|}{\textbf{Perplexity $\downarrow$}} & \multicolumn{7}{c}{\textbf{Zero-shot Accuracy $\uparrow$}} \\
      &&& \textbf{Wiki2} & \textbf{C4} & BoolQ & PIQA & Hella. & WinoG. & ARC-e & ARC-c & \textbf{Average} \\ 
    \midrule
  
    \multicolumn{1}{l}{\multirow{5}{*}{OPT-125M}} & Float16 & 16 & 27.65 & 24.60 & 55.47 & 62.02 & 31.33 & 50.19 & 39.98 & 22.86 & 43.64 \\ 
    \cmidrule(l){2-12} 
    \multicolumn{1}{c}{} & PB-LLM & 1 & 3233.63 & 1509.33 & 37.83 & 50.60 & 26.67 & 50.43 & 27.02 & 23.63 & 36.02 \\
    \multicolumn{1}{c}{} & BiLLM & 1 & 2989.53 & 1769.26 & 37.82 & 50.59 & 25.75 & 51.30 & 27.65 & 23.63 & 36.12 \\
    \multicolumn{1}{c}{} & OneBit & 1 & 39.45 & 35.58 & 61.92 & 60.01 & 27.01 & 50.43 & 35.81 & 21.84 & 42.84 \\
    \multicolumn{1}{c}{} & BinaryMoS & 1 & \textbf{36.46} & \textbf{33.13} & 61.83 & 60.17 & 27.16 & 51.38 & 36.74 & 22.95  &\textbf{43.37} \\ 
    \midrule
  
    \multicolumn{1}{l}{\multirow{5}{*}{OPT-1.3B}} & Float16 & 16 & 14.62 & 14.72 & 57.82 & 72.42 & 53.70 & 59.51 & 50.97 & 29.52 & 53.99 \\ 
    \cmidrule(l){2-12} 
    \multicolumn{1}{c}{} & PB-LLM & 1 & 272.83 & 175.42 & 62.17 & 54.24 & 27.25 & 50.27 & 27.98 & 23.72 & 40.94 \\
    \multicolumn{1}{c}{} & BiLLM & 1 & 69.45 & 63.92 & 61.92 & 59.52 & 33.81 & 49.32 & 34.38 & 22.35 & 43.55 \\
    \multicolumn{1}{c}{} & OneBit & 1 & 20.36 & 20.76 & 57.85 & 66.53 & 39.21 & 54.61 & 42.80 & 23.97 & 47.50 \\
    \multicolumn{1}{c}{} & BinaryMoS & 1 & \textbf{18.45} & \textbf{18.83} & 60.34 & 68.66 & 41.99 & 53.99 & 44.87 & 26.19 & \textbf{49.34} \\ 
    \midrule 
    \multicolumn{1}{l}{\multirow{5}{*}{LLaMA-1-7B}} & Float16 & 16 & 5.68 & 7.08 & 73.21 & 77.42 & 72.99 & 66.85 & 52.53 & 41.38 & 64.06 \\ 
    \cmidrule(l){2-12} 
    \multicolumn{1}{c}{} & PB-LLM & 1 & 198.37 & 157.35 & 60.51 & 53.53 & 27.23 & 49.17 & 27.48 & 26.02 & 40.66 \\
    \multicolumn{1}{c}{} & BiLLM & 1 & 41.66 & 48.15 & 62.23 & 58.65 & 34.64 & 51.14 & 33.08 & 25.68 & 44.24 \\
    \multicolumn{1}{c}{} & OneBit & 1 & 8.48 & 10.49 & 62.50 & 70.40 & 54.03 & 55.32 & 41.07 & 30.88 & 52.36 \\
    \multicolumn{1}{c}{} & BinaryMoS & 1 & \textbf{7.97} & \textbf{9.72} & 64.59 & 71.82 & 58.18 & 58.88 & 42.09 & 31.31 & \textbf{54.48} \\ 
    \midrule
  
    \multicolumn{1}{l}{\multirow{5}{*}{LLaMA-1-13B}} & Float16 & 16 & 5.09 & 6.61 & 68.47 & 79.05 & 76.24 & 70.17 & 59.85 & 44.54 & 66.39 \\ 
    \cmidrule(l){2-12} 
    \multicolumn{1}{c}{} & PB-LLM & 1 & 35.83 & 39.79 & 62.17 & 58.70 & 33.97 & 52.17 & 31.86 & 23.63 & 43.75 \\
    \multicolumn{1}{c}{} & BiLLM & 1 & 14.56 & 16.67 & 62.53 & 68.17 & 52.24 & 59.43 & 41.91 & 29.94 & 52.37 \\
    \multicolumn{1}{c}{} & OneBit & 1 & 7.65 & 9.56 & 63.30 & 71.98 & 60.61 & 59.43 & 42.85 & 32.42 & 55.10 \\
    \multicolumn{1}{c}{} & BinaryMoS & 1 & \textbf{7.16} & \textbf{8.81} & 63.82 & 73.88 & 64.05 & 60.93 & 44.28 & 33.11 & \textbf{56.68} \\ 
    \midrule

    \multicolumn{1}{l}{\multirow{5}{*}{LLaMA-2-7B}} & Float16 & 16 & 5.47 & 6.97 & 71.07 & 76.87 & 72.95 & 67.16 & 53.45 & 40.78 & 63.71 \\ 
    \cmidrule(l){2-12} 
    \multicolumn{1}{c}{} & PB-LLM & 1 & 76.75 & 85.92 & 62.17 & 52.82 & 26.87 & 50.11 & 26.89 & 24.31 & 40.53 \\
    \multicolumn{1}{c}{} & BiLLM & 1 & 27.72 & 36.34 & 62.14 & 59.19 & 35.18 & 53.11 & 34.22 & 26.54 & 45.06 \\
    \multicolumn{1}{c}{} & OneBit & 1 & 8.60 & 10.74 & 63.06 & 70.40 & 54.24 & 56.67 & 40.82 & 29.35 & 52.42 \\
    \multicolumn{1}{c}{} & BinaryMoS & 1 & \textbf{7.88} & \textbf{9.75} & 65.02 & 71.55 & 59.41 & 56.18 & 41.84 & 30.03 & \textbf{54.01} \\ 
    \midrule

    \multicolumn{1}{l}{\multirow{5}{*}{LLaMA-2-13B}} & Float16 & 16 & 4.88 & 6.47 & 68.99 & 79.05 & 76.62 & 69.77 & 57.95 & 44.20 & 66.10 \\ 
    \cmidrule(l){2-12} 
    \multicolumn{1}{c}{} & PB-LLM & 1 & 155.25 & 151.15 & 37.82 & 53.26 & 28.89 & 49.48 & 28.28 & 23.72 & 36.91 \\
    \multicolumn{1}{c}{} & BiLLM & 1 & 20.71 & 27.19 & 62.20 & 62.51 & 38.05 & 56.35 & 40.69 & 27.73 & 47.92 \\
    \multicolumn{1}{c}{} & OneBit & 1 & 7.56 & 9.67 & 65.66 & 71.60 & 60.07 & 56.91 & 45.76 & 31.74 & 55.29 \\
    \multicolumn{1}{c}{} & BinaryMoS & 1 & \textbf{7.08} & \textbf{8.91} & 66.12 & 73.72 & 63.80 & 58.98 & 45.71 & 33.19 & \textbf{57.09} \\ 
    %\bottomrule
    \specialrule{2.5pt}{-0.5pt}{0pt}
  \end{tabular}}
  % \vspace{-6pt}
  \vspace{-10pt}
\end{table}

\begin{table}[t]
\setlength{\tabcolsep}{3.5pt}
  \centering
  \caption{Perplexity and zero-shot accuracy results for 2-bit quantization methods and BinaryMoS.}
  \label{table:2bit_result}
  \vspace{6pt}
  \renewcommand{\arraystretch}{1.30}
  % \scalebox{.75}{
    \scalebox{.85}{
    \begin{tabular}{lc|cccccc}
    
    \specialrule{2.0pt}{0pt}{0pt}

    &&\multicolumn{6}{c}{\textbf{Perplexity $\downarrow$ \textbf{(Wikitext2)}}} \\
    % \midrule
    % \cmidrule{3-8}
    Method & Wbits & OPT-125M & OPT-1.3B & LLaMA-1-7B & LLaMA-1-13B & LLaMA-2-7B & LLaMA-2-13B \\
    \midrule
    GPTQ & 2 & 660.52 & 125.29 & 45.73 & 15.20 & 40.23 & 32.87 \\
    OmniQuant & 2 & 245.47 & 28.82 & 9.75 & 7.84 & 11.20 & 8.25 \\
    BinaryMoS & 1 & \textbf{36.46} & \textbf{18.45} & \textbf{7.97} & \textbf{7.16} & \textbf{7.88} & \textbf{7.08} \\
    \midrule

    &&\multicolumn{6}{c}{\textbf{Perplexity $\downarrow$ \textbf{(C4)}}} \\
    % \midrule
    Method & Wbits & OPT-125M & OPT-1.3B & LLaMA-1-7B & LLaMA-1-13B & LLaMA-2-7B & LLaMA-2-13B \\
    \midrule
    
    GPTQ & 2 & 213.60 & 45.43 & 27.87 & 15.15 & 31.37 & 26.23 \\
    OmniQuant & 2 & 390.30 & 33.81 & 13.01 & 10.43 & 15.46 & 11.06 \\
    BinaryMoS & 1 & \textbf{33.13} & \textbf{18.83} & \textbf{9.72} & \textbf{8.81} & \textbf{9.75} & \textbf{8.91} \\
    \midrule

    &&\multicolumn{6}{c}{\textbf{Average Zero-shot Accuracy $\uparrow$}} \\
    % \midrule
    Method & Wbits & OPT-125M & OPT-1.3B & LLaMA-1-7B & LLaMA-1-13B & LLaMA-2-7B & LLaMA-2-13B \\
    \midrule
    
    GPTQ & 2 & 37.59 & 40.36 & 43.75 & 49.65 & 43.31 & 45.03 \\
    OmniQuant & 2 & 36.54 & 46.43 & 51.58 & 56.42 & 49.54 & 54.24 \\
    BinaryMoS & 1 & \textbf{43.37} & \textbf{49.34} & \textbf{54.48} & \textbf{56.68} & \textbf{54.01} & \textbf{57.09} \\
    \specialrule{2.0pt}{0pt}{0pt}
  \end{tabular}}
  \vspace{-6pt}
\end{table}

In this section, we explore the effectiveness of the proposed mixture of scale approach in generating token-adaptive scaling factors. To accomplish this, we analyze the gating scores for scaling experts and the scaling factors derived from these scores. For this analysis, we utilize the LLaMA-1-7B model and input sequences sampled from the C4 dataset.

Figure~\ref{fig:routing_analysis} showcases the gating scores and resulting token-adaptive scaling factors for out projection of the 18th layer across tokens of the input sequence. The experimental results reveal substantial variation in the gating scores for each expert across tokens. 
As depicted in Figure~\ref{fig:routing_analysis}(b), while conventional binarization methods with static scaling factors, akin to having a single expert, offer a fixed scaling factor, the scaling experts of BinaryMoS successfully generate a diverse range of scaling factors.
This highlights the efficacy of the mixture of scale approach, which adaptively determines the scaling factor for each token, leading to a wider representation range. Consequently, we can expect that BinaryMoS effectively enhances the capacity of binarized models and improves model accuracy.

\subsection{Perplexity and Accuracy Results of Binarized Models} \label{sec:main_results}

The perplexity and zero-shot accuracy results of previous binarization methods and the proposed BinaryMoS are presented in Table~\ref{table:binarized_rsults}.
BinaryMoS consistently outperforms earlier binarization techniques across all metrics, effectively narrowing the performance disparity with their Float16 counterparts.

In particular, smaller LLMs such as OPT-125M and OPT-1.3B typically face challenges in maintaining linguistic capabilities under model compression. Previous methods like PB-LLM and BiLLM result in significant increases in perplexity, often exceeding 1000 for the OPT-125M model. While OneBit made substantial improvements, perplexity increases remained above 10. BinaryMoS, however, significantly enhances these outcomes by keeping the increase in perplexity below 10. Moreover, it boosts the accuracy of binarized models and diminishes the zero-shot accuracy gap to within 0.3\% compared to Float16 models.
The distinct advantage of BinaryMoS over previous approaches, especially OneBit, lies in its use of scaling experts. This evaluation underlines the efficacy of the BinaryMoS with mixture of scales approach.

\subsection{Comparison between BinaryMoS and 2-bit Quantization} \label{sec:2bit_compare}

Since BinaryMoS consistently outperforms other binarization methods, we proceed to compare it with conventional 2-bit quantization techniques, GPTQ and OmniQuant. While these two approaches entail lower calibration overhead for quantization due to their use of the PTQ approach, they differ in their quantization methods. GPTQ and OmniQuant utilize a group-wise quantization approach, employing groups of 128 weights to finely quantize parameters and minimize quantization errors. Consequently, the memory demand during inference for these methods is more than double that of BinaryMoS.
The comparison results, presented in Table~\ref{table:2bit_result}, reveal that BinaryMoS even outperforms these 2-bit quantization methods, despite its lower memory requirement during inference. This once again underscores the effectiveness of integrating scaling experts.

\section{Discussion and Future Work}

BinaryMoS significantly improves the accuracy of binarized LLMs by increasing their representational capability with mixture of scales. This MoS approach holds promise for extension to multi-bit quantization, as multi-bit quantization techniques also involve scaling factors for regulating quantization step size. However, in this paper, our study does not delve into the effectiveness of the mixture of scales on multi-bit quanization schemes, leaving this avenue for future exploration.

Though BinaryMoS adopts the concept of MoE, it does not fully leverage advanced training techniques established in the field of MoE~\cite{moe, switch_trans, stmoe}. These advanced methods optimize routing functions and balance token assignments among experts, thereby enhancing MoE model accuracy. Thus, investigating these training techniques is another topic for future research.

\section{Conclusion}
This paper introduces BinaryMoS, a novel binarization technique designed to enhance the representation capability of binarized LLMs while preserving the fundamental advantage of binarization—low memory usage. BinaryMoS adopts the mixture of scale approach to dynamically adjust the scaling factors of binary weight values in a token-adaptive manner. Given that scaling factors play a crucial role in reducing binarization error and occupy a small portion of binarized models, this approach effectively mitigates information loss associated with binarization with minimal memory overhead. Our experimental findings demonstrate that BinaryMoS surpasses existing binarization approaches and even outperforms 2-bit quantization methods in both perplexity and zero-shot tasks.

\section*{Acknowledgements}
This work was supported in part by Institute of Information \& communications Technology Planning \& Evaluation (IITP) grant funded by the Korea government (MSIT) (No. 2021-0-00105: Development of model compression framework for scalable on-device AI computing on Edge applications , IITP-2023-RS-2023-00256081: artificial intelligence semiconductor support program to nurture the best talents, No. 2021-0-01343: Artificial Intelligence Graduate School Program (Seoul National University) , No.2021-0-02068: Artificial Intelligence Innovation Hub), in part by the Samsung Research Funding Center under Project SRFC-TC1603-53, and BK21 FOUR program at Seoul National University.

%%%%%%%%%%%%%%%%%%%%%%%%%%%%%%%%%%%%%%%%%%%%%%%%%%%%%%%%%%%%

%%%%%%%%%%%%%%%%%%%%%%%%%%%%%%%%%%%%%%%%%%%%%%%%%%%%%%%%%%%%
\newpage
\appendix

\section{Appendix}

\subsection{Ablation Study on Datasets} \label{sec:dataset}

 To determine the optimal dataset for training binarized models, we conduct a comparative analysis using various training datasets, as summarized in Table~\ref{table:dataset}. The results indicate that models trained solely on the WikiText2 dataset, due to its relatively small dataset size, tend to exhibit overfitting tendencies and struggle to generalize to other datasets. While these models demonstrate considerable perplexity improvement on the WikiText2 evaluation, their perplexity on the C4 dataset and zero-shot accuracy is notably poor.
Conversely, models trained exclusively on the C4 dataset perform well across a wide range of tasks, except for the evaluation on WikiText2. Following the approach of previous research~\cite{llm_qat}, we also experiment with a generated dataset synthesized using the LLaMA-1-7B model. Although this dataset generally performs satisfactorily across various language modeling tasks, its performance lags behind that of the C4 dataset.
Therefore, to enhance overall model performance, we opt to train the models on a mixed dataset comprising both C4 and WikiText2. Moreover, the accessibility of both C4 and WikiText2 as open-source datasets further facilitates their adoption for training purposes.

 \begin{table}[h]
 \setlength{\tabcolsep}{3.5pt}
   \centering
   \caption{
   Evaluation of binarized LLaMA-1-7B model trained with various training datasets. We train the model on a subset of the dataset with the same training step. $\dagger$: Generated dataset synthesized by LLaMA-1-7B model. $\ddagger$: Mixed dataset of Wikitext2 and C4.
   }
   \label{table:dataset}
   \vspace{6pt}
   \renewcommand{\arraystretch}{1.20}
   \scalebox{0.95}{
     \begin{tabular}{c|cc|ccccccc}
     \toprule
     \multirow{2}{*}{\hspace{-0.15cm}\begin{tabular}{c}\textbf{Training} \\\textbf{Dataset}\end{tabular}} & \multicolumn{2}{c|}{\textbf{Perplexity $\downarrow$}} & \multicolumn{7}{c}{\textbf{Zero-shot Accuracy $\uparrow$}} \\ 
     & \textbf{Wiki2} & \textbf{C4} & \textbf{BoolQ} & \textbf{PIQA} & \textbf{Hella.} & \textbf{WinoG.} & \textbf{ARC-e} & \textbf{ARC-c} & \textbf{Avg} \\ 
     \midrule
     Generated $\dagger$ & 12.54 & 13.04 & 60.51 & 66.10 & 45.91 & 54.69 & 41.41 & 27.90 & 49.42 \\
     Wiki2 & 9.65 & 28.61 & 57.95 & 57.67 & 36.78 & 54.45 & 38.46 & 26.62 & 45.32 \\
     C4 & 13.76 & 11.97 & 60.33 & 67.79 & 49.69 & 53.74 & 39.26 & 29.52 & 50.06 \\
     Mixed $\ddagger$& 8.92 & 11.85 & 60.51 & 67.46 & 49.95 & 55.24 & 41.16 & 29.35 & 50.61 \\
     \bottomrule
   \end{tabular}}
   \vspace{-6pt}
 \end{table}

\subsection{Latency Measurement} \label{sec:latency}
To assess the latency of GEMV operation for our BinaryMoS, we have evaluated the latency of previous binarized models and the BinaryMoS by developing appropriate CUDA kernels for 1-bit matrix multiplication, modifying the CUDA kernel for multi-bit matrix multiplication~\cite{quick}. Additionally, we further customize the CUDA kernel of BinaryMoS to fuse scaling experts and routing operations on top of the 1-bit matrix multiplication CUDA kernel. We measure the latency of the linear layers in LLaMA-7B and LLaMA-13B with batch size of 1 and results are presented in Table~\ref{table:latency}. All experiments are conducted on NVIDIA A6000 GPUs.

Previous methods like PB-LLM and BiLLM require extra matrix multiplications, making them very slow. OneBit, which employs the simplest binarization scheme, achieves significant improvement over the original Float16 model and shows the minimum latency. Meanwhile, our BinaryMoS introduces additional operations for processing scaling experts, which require far fewer operations compared to matrix multiplication. Consequently, BinaryMoS also shows similar latency results to OneBit. This demonstrates that the multi-scaling factor module in BinaryMoS improves performance in terms of perplexity and zero-shot accuracy with minimal overhead to latency.

 \begin{table}[h]
 \setlength{\tabcolsep}{3.5pt}
   \centering
   \caption{
   Latency ($\mu$sec) of linear layer in LLaMA-1/2-7B and LLaMA-1/2-13B.
   }
   \label{table:latency}
   \vspace{6pt}
   \renewcommand{\arraystretch}{1.20}
   \scalebox{0.88}{
     \begin{tabular}{l|r|r|r|r|r|r}
     \toprule
     \textbf{Model Config} & \multicolumn{3}{c|}{\textbf{LLaMA-1/2-7B}} & \multicolumn{3}{c}{\textbf{LLaMA-1/2-13B}} \\ 
     \midrule
     \textbf{Weight Size} & \small \textbf{4096 $\times$ 4096} & \small \textbf{4096 $\times$ 11008} & \small \textbf{11008 $\times$ 4096} & \small \textbf{5120 $\times$ 5120} & \small \textbf{5120 $\times$ 13824} & \small \textbf{13824 $\times$ 5120} \\ 
     \midrule
     Float16 & 68.2 $\:\:\:\:\:\:$& 151.7 $\:\:\:\:\:\:$& 143.5 $\:\:\:\:\:\:$& 95.6 $\:\:\:\:\:\:$& 224.1 $\:\:\:\:\:\:$& 213.6 $\:\:\:\:\:\:$\\
     PB-LLM & 96.1 $\:\:\:\:\:\:$& 177.5 $\:\:\:\:\:\:$& 168.3 $\:\:\:\:\:\:$& 122.7 $\:\:\:\:\:\:$& 243.7 $\:\:\:\:\:\:$&$\:\:\:\:\:\:$ 234.7 $\:\:\:\:\:\:$\\
     BiLLM & 87.1 $\:\:\:\:\:\:$& 96.4 $\:\:\:\:\:\:$& 104.2 $\:\:\:\:\:\:$& 95.2 $\:\:\:\:\:\:$& 124.2 $\:\:\:\:\:\:$& 131.0 $\:\:\:\:\:\:$\\
     OneBit & 32.7 $\:\:\:\:\:\:$& 33.7 $\:\:\:\:\:\:$& 34.9 $\:\:\:\:\:\:$& 33.4 $\:\:\:\:\:\:$& 41.4 $\:\:\:\:\:\:$& 42.6 $\:\:\:\:\:\:$\\
     BinaryMoS & 34.5 $\:\:\:\:\:\:$& 36.9 $\:\:\:\:\:\:$& 37.0 $\:\:\:\:\:\:$& 35.6 $\:\:\:\:\:\:$& 43.4 $\:\:\:\:\:\:$& 44.5  $\:\:\:\:\:\:$\\
     \bottomrule
   \end{tabular}}
   \vspace{-6pt}
 \end{table}

\subsection{Experimental Results for LLaMA-1-30B}
Table~\ref{table:llama_30b} provides further experimental results on the LLaMA-1-30B model. In line with the trends observed in Table~\ref{table:binarized_rsults}, BinaryMoS consistently surpasses other binarization approaches for this 30B model. This assessment highlights the effectiveness of BinaryMoS for large-scale LLMs.

\begin{table}[h]
\setlength{\tabcolsep}{3.8pt}
  \centering
  \caption{Perplexity and zero-shot accuracy results of Float16 and binarized LLMs for LLaMA-1-30B}
  \label{table:llama_30b}
  \vspace{6pt}
  \renewcommand{\arraystretch}{1.27}
  % \scalebox{.75}{
  \scalebox{.83}{
    \begin{tabular}{llc|cc|cccccc|c}
    \toprule
    %\toprule
    \multirow{2}{*}{\textbf{Model}} & \multirow{2}{*}{\textbf{Method}} & \multirow{2}{*}{\textbf{Wbits}} & \multicolumn{2}{c|}{\textbf{Perplexity $\downarrow$}} & \multicolumn{7}{c}{\textbf{Zero-shot Accuracy $\uparrow$}} \\
      &&& \textbf{Wiki2} & \textbf{C4} & BoolQ & PIQA & Hella. & WinoG. & ARC-e & ARC-c & \textbf{Average} \\ 
    \midrule
  
    \multicolumn{1}{l}{\multirow{5}{*}{LLaMA-1-30B}} & Float16 & 16 & 4.10 & 5.98 & 68.37 & 80.08 & 79.22 & 72.69 & 58.96 & 45.39 & 67.45 \\ 
    \cmidrule(l){2-12} 
    \multicolumn{1}{c}{} & PB-LLM & 1 & 32.24 & 34.59 & 62.18 & 60.22 & 33.75 & 53.91 & 33.50 & 24.40 & 44.66 \\
    \multicolumn{1}{c}{} & BiLLM & 1 & 10.10 & 12.28 & 62.39 & 71.38 & 59.49 & 66.54 & 44.95 & 33.19 & 56.32 \\
    \multicolumn{1}{c}{} & BinaryMoS & 1 & \textbf{6.63} & \textbf{8.36} & 67.76 & 75.62 & 67.38 & 64.09 & 49.62 & 35.07  &\textbf{59.92} \\ 
    \bottomrule
  \end{tabular}}
  % \vspace{-6pt}
  \vspace{-10pt}
\end{table}

\subsection{Generation Quality}
\begin{figure}[h]
    \centering
    \includegraphics[width=1.0\linewidth]{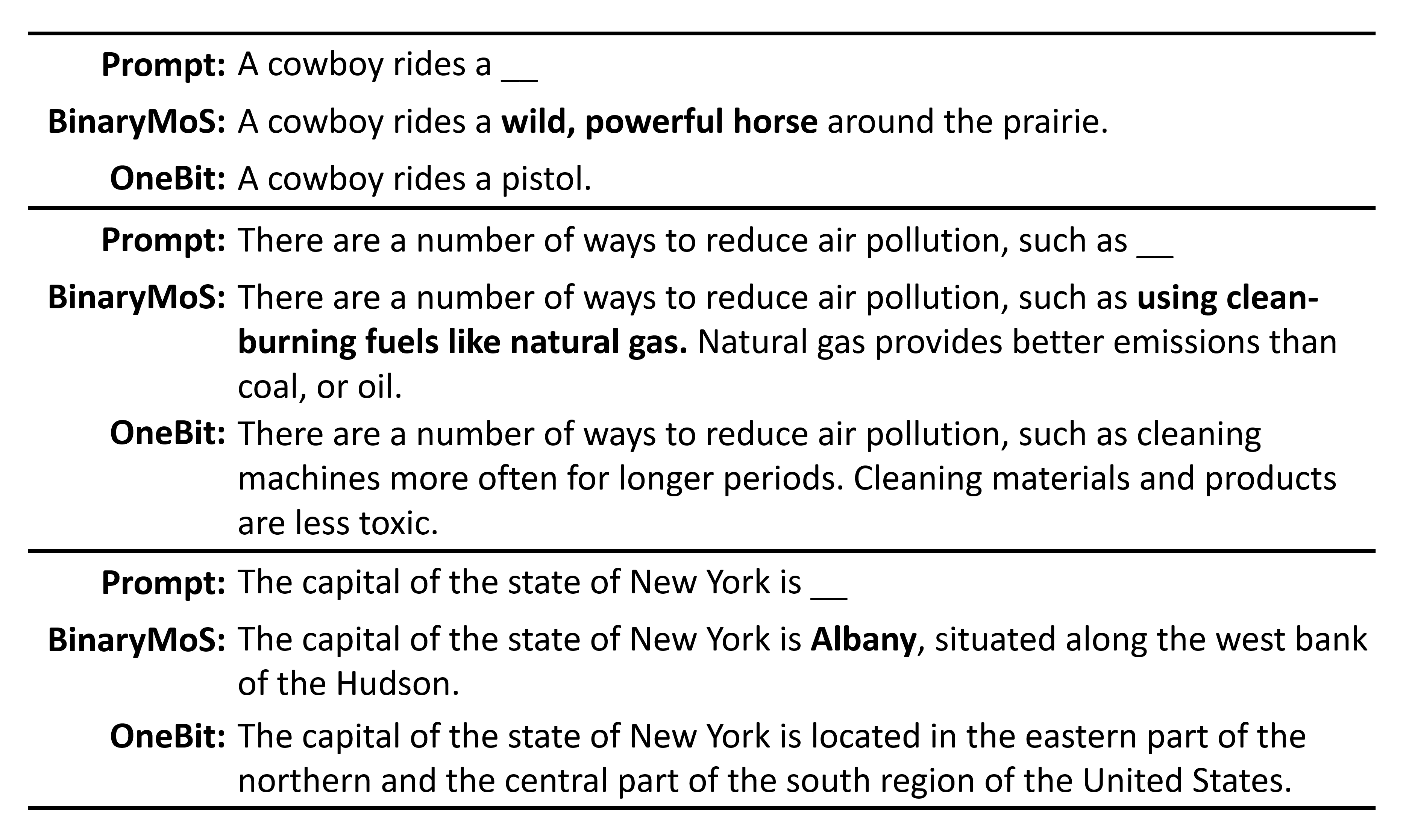}
    % \vspace{-12pt}
        \vspace{-18pt}
    \caption{
        Comparison of generation quality on the LLaMA-1-13B models with BinaryMoS and OneBit.
    }
    \vspace{-10pt}
    \label{fig:generation_quality}
\end{figure}

Figure~\ref{fig:generation_quality} compares the generation quality between BinaryMoS and OneBit on LLaMA-1-13B model. BinaryMoS can generate contextually proper answers, whereas OneBit fails to generate correct answers. These results demonstrate that the BinaryMoS processes each token with token-adaptive scaling factors which contain contextual information and the improvement of BinaryMoS over OneBit is substantial enough to make binarized LLMs more applicable in practice.

\vspace{-10pt}

\subsection{Limitations} \label{sec:limitation}
While BinaryMoS effectively enhances the linguistic capabilities of binarized LLMs, its application to extremely large models such as LLaMA-2-70B poses challenging due to the high training cost inherent to QAT-based strategies. Hence, to facilitate the use of BinaryMoS in such large-scale models, it may be necessary to adapt it into a PTQ-based approach or to integrate it with parameter-efficient training strategies in future work.
Additionally, it is important to note that, despite the advancements brought about by by BinaryMoS, the reduction in linguistic performance of binarized LLMs coampred to their Float16 counterparts remains substantial. Consequently, to make LLM binarization practical for real-world applications, further advancements in binarization techniques are required.

\end{document}